\pdfoutput=1

\documentclass[11pt]{article}

\usepackage[final]{acl}
\usepackage{colortbl}
\usepackage{ulem}
\definecolor{shadecolor}{gray}{.9}
\definecolor{Gray}{gray}{0.9}
\usepackage{times}
\usepackage{latexsym}
\usepackage{booktabs}
\usepackage[T1]{fontenc}

\usepackage[utf8]{inputenc}

\usepackage{microtype}
\usepackage[namelimits]{amsmath}
\usepackage{inconsolata}

\usepackage{graphicx}
\usepackage{multirow}
\usepackage{subcaption}
\usepackage{float}
\usepackage{enumitem}
\usepackage{verbatim}
\usepackage{colortbl}
\usepackage{mdframed}
\usepackage{adjustbox}
\usepackage{tabularx}
\usepackage{enumitem}
\usepackage{tcolorbox}
\tcbuselibrary{skins, breakable}  
\newcommand{\our}[1]{\textsc{NL-Debugging}}
\newcommand{\task}[1]{Debugging}

\title{NL-Debugging: Exploiting Natural Language as an Intermediate Representation for Code Debugging}

\author{
Weiming Zhang$^{1}$\thanks{Equal Contribution.}, 
Qingyao Li$^{1}$\footnotemark[1], 
Xinyi Dai$^{1}$, Jizheng Chen$^{1}$, Kounianhua Du$^{1}$, \\
\bf Weiwen Liu$^{1}$\thanks{Corresponding author.}, Yasheng Wang$^{2}$, Ruiming Tang$^{2}$, Yong Yu$^{1}$, Weinan Zhang$^{1}$ \\
$^{1}$Shanghai Jiao Tong University, $^{2}$Huawei Noah’s Ark Lab \\
Shanghai, China \\
\texttt{\{WeimingZhang\_2020, ly890306, wwliu\}@sjtu.edu.cn}
}

\begin{document}
\maketitle
\begin{abstract}
Debugging is a critical aspect of LLM's coding ability. Early debugging efforts primarily focused on code-level analysis, which often falls short when addressing complex programming errors that require a deeper understanding of algorithmic logic. Recent advancements in large language models (LLMs) have shifted attention toward leveraging natural language reasoning to enhance code-related tasks. However, two fundamental questions remain unanswered: What type of natural language format is most effective for debugging tasks? And what specific benefits does natural language reasoning bring to the debugging process? In this paper, we introduce \our{}, a novel framework that employs natural language as an intermediate representation to improve code debugging. By debugging at a natural language level, we demonstrate that \our{} outperforms traditional debugging methods and enables a broader modification space through direct refinement guided by execution feedback. Our findings highlight the potential of natural language reasoning to advance automated code debugging and address complex programming challenges.

\end{abstract}
\section{Introduction}


Code generation is a challenging task that requires strong reasoning abilities and a deep understanding of programming languages. While large language models (LLMs) demonstrate potential in this domain~\cite{achiam2023gpt,  zhu2024deepseek, li2023starcoder}, they often struggle to produce fully correct code implementations in one attempt~\cite{liu2024your, dou2024s}. This limitation has driven recent research toward iterative refinement approaches for code generation~\cite{chen2023teaching, zhong2024ldb, chen2025debatecoder}, where the debugging process—transforming flawed code into correct implementations—has emerged as a critical research focus.

Early debugging methodologies primarily relied on execution feedback for code-level analysis~\cite{zhang2023self,chen2023teaching, zhong2024ldb}. Although effective for identifying syntax errors or fundamental logical flaws, these approaches show limited capability in detecting deeper algorithmic design issues~\cite{finder2010evaluating, tian2024debugbench}. Inspired by the success of natural language reasoning in LLMs~\cite{jaech2024openai, guo2025deepseek, qin2024o1}, recent advancements have shifted focus to natural language-based code optimization~\cite{wang2024planning, li2024rethinkmcts, li2024codetree}. This trend reflects a growing recognition of the potential for natural language to serve as a powerful medium for guiding and improving code-related tasks.



Despite promising progress, two fundamental research questions about natural language reasoning in debugging remain unanswered: (1) What type of natural language format is most effective for debugging tasks? Existing work often assumes a specific natural language format, such as pseudocode~\cite{zhang2024o1}, thought points~\cite{wang2024planning}, or sketches, without exploring why these formats are effective or whether alternatives might perform better. (2) What specific benefits does natural language reasoning bring to the debugging process? While empirical results validate the utility of natural language reasoning, its underlying mechanisms for improving debugging success remain unclear.

In this paper, we propose \textbf{N}atural \textbf{L}anguage for Code \textbf{Debugging} (\our{}), a framework that debugs code at the natural language level. The framework contains three key phases: Backtranslation, Natural Language Refinement, and Regeneration. Specifically, the framework first translates buggy code into natural language representations. It then debugs these representations in the natural language space, producing a refined natural language version of the code. Finally, the corrected code is regenerated based on the refined natural language representation. 


The main contributions of this paper are novel methods to debug code at the natural language level and findings that shed light on how natural language helps LLMs debug as an intermediate representation. We conduct extensive experiments using our framework to systematically investigate the limitations, underlying principles, and key factors for effectively leveraging natural language as an intermediate representation in code debugging. And we summarize key findings as follows:


    
\paragraph{Sketch as Intermediate Natural Language Representation Brings Substantial Debugging Performance Gains.} Comparing to other debugging methods, Sketch-based NL-debugging significantly improves in a large margin for debugging in natural language space.
\paragraph{Natural Language Debugging Enhances Modification Space and Diversity.} The effectiveness of natural language reasoning lies in its ability to provide a broader modification space, increasing diversity and enabling more effective corrections, especially for complex algorithmic errors.
\paragraph{Feedback and Step-wise Analysis Drive NL-Debugging Efficacy.} Combining execution feedback with iterative natural language refinement yields enhanced debugging results, supporting more structured and iterative improvements.

Our findings highlight the critical role of natural language in bridging the gap between LLM-based debugging and human-like reasoning, paving the way for future research on enhancing code debugging through semantic refinement.

\section{Related Work}

\subsection{LLMs for Code Debugging}
Code debugging is essential in software development, focusing on the automatic correction of code bugs~\cite{gupta2020synthesize, yasunaga2021break}. Two primary approaches utilize large language models (LLMs) for this purpose. The first approach involves training LLMs on task-specific datasets~\cite{huang2023empirical, jiang2024training, zheng2024opencodeinterpreter, kumar2024training}. However, its effectiveness is constrained by the quality and scope of the training data, which directly impacts the model's ability to handle various bugs. The second approach capitalizes on the reasoning capabilities of pretrained LLMs, which analyze buggy code and suggest fixes based on prior knowledge and execution feedback~\cite{zhang2023self,madaan2024self, chen2023teaching, zhong2024ldb, hu2024leveraging}. Recent advancements have explored various techniques that harness these reasoning abilities. For instance, Self-Debugging~\cite{chen2023teaching} prompts LLMs to explain or dry-run generated programs, similar to rubber duck debugging. LDB~\cite{zhong2024ldb} segments programs into basic blocks and tracks variable values during runtime to verify correctness against task descriptions. MGDB~\cite{shi2024code} decomposes problematic code into a hierarchical tree structure of subfunctions for bottom-up analysis.

While these methods have significantly advanced automatic program repair, they primarily rely on local code analysis, limiting their effectiveness in complex programming scenarios~\cite{xia2023automated, hossain2024deep}. To address these challenges, we investigate approaches that utilize natural language as an intermediary to facilitate more comprehensive and scalable debugging processes. This strategy aims to overcome the limitations of traditional code-level analysis and tackle more intricate debugging tasks.

\begin{figure*}[h]
    \centering
    \includegraphics[width=\textwidth]{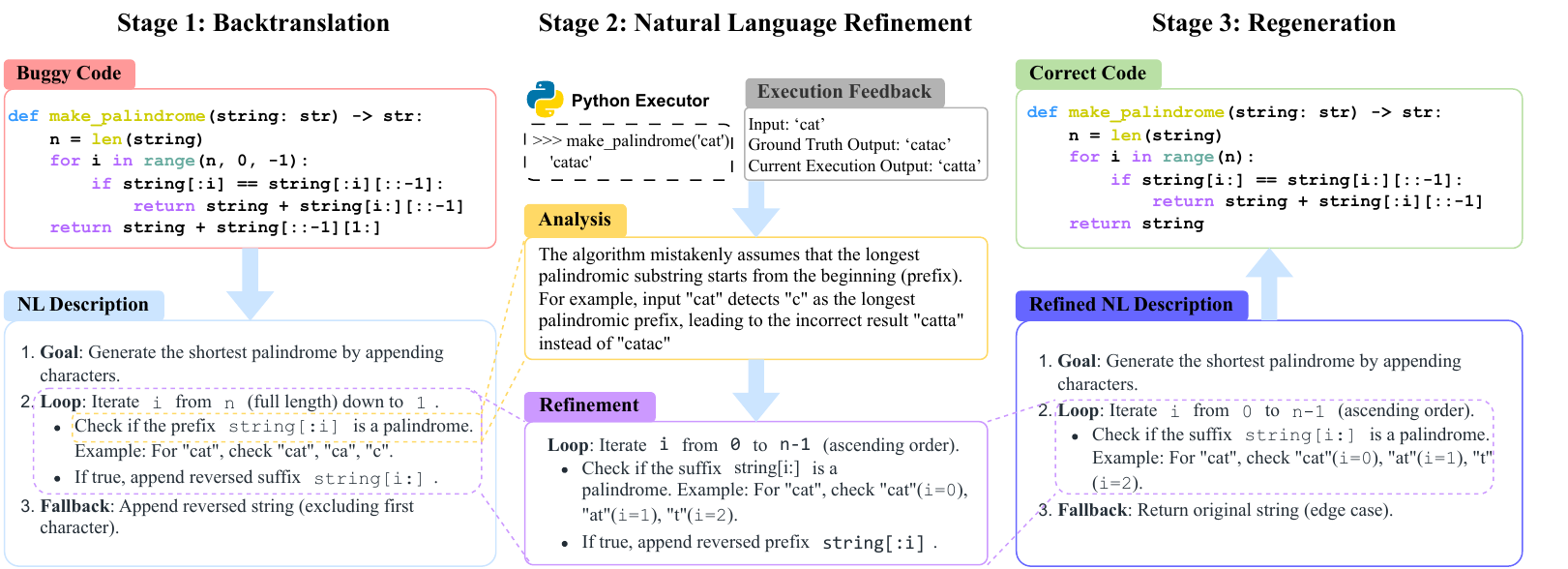}
    \caption{The \our{} framework. This iterative process includes backtranslation, refinement, and regeneration, ultimately improving debugging efficiency by utilizing natural language reasoning.}
    \label{fig:overview}
\end{figure*}

\subsection{Code Generation with Natural Language Reasoning}
Recent advances in LLM reasoning have demonstrated strong potential in handling complex tasks, with natural language playing a central role~\cite{huang2022towards, yang2025agentnet}. Initially, Chain-of-Thought (CoT) methods~\cite{wei2022chain, wang2022self} organized reasoning into step-by-step natural language explanations. Building on this, plan-based approaches like Tree-of-Thought (ToT)~\cite{yao2024tree} and Graph-of-Thought~\cite{besta2024graph} introduced structured search spaces to explore multiple reasoning paths, expanding the solution space. More recent methods such as o1~\cite{jaech2024openai}, STaR~\cite{zelikman2022star}, and Iter-MCTS~\cite{xie2024monte} focus on iterative refinement, enabling models to progressively improve reasoning~\cite{li2025codeprm, zhu-etal-2025-retrieval}.

Recent research has started incorporating natural language reasoning explicitly into code generation. For instance, explorations like RethinkMCTS~\cite{li2024rethinkmcts} and CodeTree~\cite{li2024codetree} employ tree search techniques within the natural language space, iteratively refining reasoning steps to improve code quality. However, many existing methods assume a fixed natural language format and overlook how different structures can impact reasoning effectiveness. This motivates our investigation into how varying natural language representations can better support LLM-driven code generation and debugging.

\section{Method}

We propose \our{} that enhances code debugging by utilizing natural language as an intermediate representation. The workflow of \our{} is illustrated in Figure~\ref{fig:overview}. In the \textbf{Backtranslation} stage, the buggy code is converted into a natural language version that captures its underlying logic. Next, in the \textbf{Natural Language Refinement} stage, execution feedback is used to analyze the natural language representation, identifying discrepancies between the intended behavior and the actual execution. Based on this analysis, the erroneous representation is modified to correct identified errors. Finally, in the \textbf{Regeneration} stage, the refined natural language representation is translated back into executable code, generating a new implementation to resolve the identified issues. This iterative process continues until the program passes all visible test cases or reaches a predefined iteration limit, with the final solution evaluated against hidden test cases to assess its generalizability.

\subsection{Backtranslation}
The first step in our framework involves back-translating the buggy code into a natural language form. This transformation captures the intent and structure of the code. We facilitate semantic reasoning and high-level debugging by representing the program logic in natural language. This natural language representation can take various forms, such as sketches, pseudocode, or key thought points, and we present the details in Appendix~\ref{pt:prompts}.

As illustrated in Figure~\ref{fig:overview}, this process translates a function implementation into a natural language format that provides an interpretable description of how the function is expected to behave. This representation allows us to concentrate on the overall problem-solving strategy rather than getting bogged down in syntax-specific details. Detailed information on backtranslating to these different natural language forms can be found in Appendix~\ref{app:different}.

\subsection{Natural Language Refinement}

After generating the initial natural language version of the buggy code, the next step is refinement, which improves the representation based on execution feedback. This process consists of three key stages:

1. \textbf{Execution Feedback Collection:} The buggy code is executed against a set of test cases, and runtime feedback is systematically collected by comparing the predicted output against the ground truth. As shown in Figure~\ref{fig:overview}, each feedback explicitly contains input, ground truth output, and current execution output, allowing for precise identification of discrepancies (e.g., an erroneous character duplication on line 3).

2. \textbf{Problem Analysis and Reasoning:} Based on the problem description, the buggy code with natural language representation, and the execution feedback, we analyze the nature of the issue and reason about its root causes. This involves logical deductions about the errors in the code and how they relate to the problem requirements.

3. \textbf{Refinement of the Natural Language Sketch:} After completing the analysis, we refine the natural language representation to produce an updated version, which better aligns with the correct solution logic.

By integrating execution feedback and problem analysis, the natural language representation evolves, improving its alignment with the correct solution and addressing previously identified logical flaws.

\subsection{Regeneration}

After refinement, we regenerate the corrected code using the refined natural language representation containing the correct implementation logic. The process involves translating the refined natural language representation back into executable code. 

The framework iteratively runs the backtranslation, refinement, and regeneration steps until the program passes all visible test cases or reaches the maximum allowed debugging iterations. The finalized solution is then evaluated against hidden test cases to assess its robustness. If the solution passes the hidden test cases, it is considered correct.

\section{Experiment}
In this section, we conduct a series of experiments to answer the following research questions (RQs):
\begin{itemize}[leftmargin=27pt]
    \item[\textbf{RQ1}] How does \our{} framework perform against other code debugging methods?
    \item[\textbf{RQ2}] What type of natural language format is most effective for debugging tasks?
    \item[\textbf{RQ3}] Why is natural language beneficial for code debugging process?
    \item[\textbf{RQ4}] How does natural language work better with code execution feedback?
    \item[\textbf{RQ5}]Is o1-like long-CoT beneficial for \our{}?
\end{itemize}

\subsection{Experiment Settings}

\subsubsection{Datasets}
We evaluate our \our{} framework on two widely used code datasets: APPS~\cite{hendrycks2021measuring} and Codeforces~\citep{codeforcesweb}. The APPS dataset consists of three difficulty levels—introductory, interview, and competition. From each difficulty level, we selected 100 problems to ensure a balanced assessment. The Codeforces dataset includes problems from online programming contests categorized by "ratings". We chose problems with ratings of 1200, 1500, and 1800 and selected 100 problems per rating level. Within each difficulty and rating category, problems were randomly selected to ensure balanced coverage, following  \citet{zhang2023planning, islam2024mapcoder}. We choose the first 100 problems per difficulty or rating level to maintain randomness.

\subsubsection{Metrics}
We use two common evaluation metrics to assess code correctness: pass rate and pass@1 following~\citet{zhang2023planning}. The pass rate is the average percentage of private test cases the generated programs pass across all problems. At the same time, pass@1 is the percentage of problems where the generated programs pass all private test cases, the standard metric in code-related tasks~\cite{chen2021evaluating,dong2023codescore}.

\begin{table*}[t]
\centering

\caption{Code debugging performance comparison of \our{} and other code debugging methods on APPS and Codeforces. We report pass rate and pass@1 on both datasets and all difficulties.}
\label{tab:my-table}
\resizebox{\textwidth}{!}{%
\begin{tabular}{@{}clcccccccccccc@{}}
\toprule
\multirow{3}{*}{\textbf{Arch}} &
  \multicolumn{1}{c}{\multirow{3}{*}{\textbf{Method}}} &
  \multicolumn{6}{c}{\textbf{APPS}} &
  \multicolumn{6}{c}{\textbf{Codeforces}} \\
 &
  \multicolumn{1}{c}{} &
  \multicolumn{3}{c}{\textbf{Pass Rate(\%)}} &
  \multicolumn{3}{c}{\textbf{Pass@1(\%)}} &
  \multicolumn{3}{c}{\textbf{Pass Rate(\%)}} &
  \multicolumn{3}{c}{\textbf{Pass@1(\%)}} \\ \cmidrule(l){3-14} 
 &
  \multicolumn{1}{c}{} &
  \textbf{Intro.} &
  \textbf{Inter.} &
  \textbf{Comp.} &
  \textbf{Intro.} &
  \textbf{Inter.} &
  \textbf{Comp.} &
  \textbf{1200} &
  \textbf{1500} &
  \textbf{1800} &
  \textbf{1200} &
  \textbf{1500} &
  \textbf{1800} \\ \midrule
\multirow{8}{*}{\textbf{GPT-4o-mini}} &
  Base(w/o debugging) &
  56.45 &
  54.57 &
  34.67 &
  35 &
  28 &
  16 &
  64.06 &
  47.60 &
  36.60 &
  42 &
  23 &
  15 \\
 &
  Self-Edit &
  64.85 &
  61.14 &
  40.00 &
  47 &
  34 &
  20 &
  71.32 &
  55.12 &
  43.67 &
  52 &
  30 &
  19 \\
 &
  Self-Debugging(Expl.) &
  66.78 &
  68.39 &
  43.17 &
  48 &
  44 &
  22 &
  76.01 &
  61.72 &
  45.81 &
  60 &
  37 &
  23 \\
 &
  Self-Debugging(Trace) &
  65.63 &
  64.57 &
  36.33 &
  47 &
  42 &
  19 &
  76.09 &
  60.78 &
  47.68 &
  59 &
  37 &
  22 \\
 &
  LDB &
  64.53 &
  63.11 &
  38.85 &
  47 &
  38 &
  19 &
  73.91 &
  56.42 &
  43.58 &
  53 &
  32 &
  20 \\
 &
  MGDB &
  61.13 &
  59.57 &
  41.33 &
  45 &
  35 &
  20 &
  69.13 &
  54.99 &
  42.33 &
  48 &
  28 &
  17 \\
 &
  Reflexion &
  64.48 &
  62.08 &
  42.50 &
  45 &
  36 &
  21 &
  74.21 &
  55.47 &
  41.89 &
  56 &
  29 &
  17 \\
  \rowcolor{Gray}
 &
  \our{} &
  \textbf{71.36} &
  \textbf{69.96} &
  \textbf{44.17} &
  \textbf{51} &
  \textbf{48} &
  \textbf{23} &
  \textbf{79.57} &
  \textbf{64.01} &
  \textbf{48.69} &
  \textbf{63} &
  \textbf{41} &
  \textbf{25} \\ \midrule
\multirow{8}{*}{\textbf{Claude3.5-Sonnet}} &
  Base(w/o debugging) &
  60.01 &
  61.47 &
  45.33 &
  41 &
  36 &
  30 &
  68.59 &
  56.33 &
  42.06 &
  56 &
  33 &
  24 \\
 &
  Self-Edit &
  70.61 &
  76.57 &
  56.83 &
  51 &
  57 &
  40 &
  81.83 &
  66.25 &
  54.60 &
  73 &
  44 &
  40 \\
 &
  Self-Debugging(Expl.) &
  71.43 &
  75.12 &
  59.17 &
  56 &
  61 &
  42 &
  83.32 &
  70.92 &
  59.11 &
  75 &
  53 &
  44 \\
 &
  Self-Debugging(Trace) &
  73.04 &
  75.23 &
  56.83 &
  58 &
  60 &
  42 &
  84.29 &
  71.00 &
  58.32 &
  76 &
  51 &
  44 \\
 &
  LDB &
  70.29 &
  73.20 &
  58.33 &
  55 &
  53 &
  42 &
  83.93 &
  69.94 &
  60.75 &
  73 &
  53 &
  46 \\
 &
  MGDB &
  70.27 &
  68.45 &
  55.67 &
  53 &
  44 &
  40 &
  77.35 &
  65.10 &
  55.14 &
  66 &
  43 &
  39 \\
 &
  Reflexion &
  75.99 &
  74.42 &
  54.17 &
  55 &
  55 &
  39 &
  83.58 &
  72.01 &
  59.78 &
  74 &
  56 &
  44 \\
  \rowcolor{Gray}
 &
  \our{} &
  \textbf{77.98} &
  \textbf{76.72} &
  \textbf{59.67} &
  \textbf{62} &
  \textbf{62} &
  \textbf{44} &
  \textbf{85.82} &
  \textbf{74.16} &
  \textbf{61.28} &
  \textbf{79} &
  \textbf{58} &
  \textbf{47} \\ \midrule
\multirow{8}{*}{\textbf{DeepSeek-Coder-V2-Lite}} &
  Base(w/o debugging) &
  50.79 &
  45.42 &
  21.33 &
  33 &
  24 &
  6 &
  52.64 &
  38.33 &
  19.75 &
  33 &
  17 &
  5 \\
 &
  Self-Edit &
  53.30 &
  47.51 &
  22.83 &
  34 &
  25 &
  7 &
  54.40 &
  42.37 &
  23.27 &
  36 &
  21 &
  8 \\
 &
  Self-Debugging(Expl.) &
  59.63 &
  57.43 &
  25.00 &
  \textbf{41} &
  30 &
  10 &
  58.39 &
  45.08 &
  29.78 &
  41 &
  22 &
  7 \\
 &
  Self-Debugging(Trace) &
  55.90 &
  49.83 &
  23.83 &
  \textbf{41} &
  27 &
  7 &
  55.77 &
  43.89 &
  25.27 &
  37 &
  22 &
  7 \\
 &
  LDB &
  56.78 &
  49.39 &
  19.67 &
  39 &
  27 &
  7 &
  54.40 &
  40.68 &
  24.65 &
  35 &
  20 &
  7 \\
 &
  MGDB &
  56.29 &
  54.78 &
  26.50 &
  36 &
  29 &
  8 &
  57.28 &
  42.24 &
  24.13 &
  43 &
  21 &
  6 \\
 &
  Reflexion &
  54.01 &
  49.82 &
  20.83 &
  35 &
  26 &
  7 &
  56.17 &
  42.18 &
  23.74 &
  38 &
  20 &
  6 \\
  \rowcolor{Gray}
 &
  \our{} &
  \textbf{62.23} &
  \textbf{59.23} &
  \textbf{27.60} &
  40 &
  \textbf{32} &
  \textbf{12} &
  \textbf{63.65} &
  \textbf{50.92} &
  \textbf{32.80} &
  \textbf{48} &
  \textbf{27} &
  \textbf{12} \\ \bottomrule
\end{tabular}%
}
\end{table*}

\subsubsection{Baselines}
To investigate how \our{} performs, we compare it with a series of code debugging methods, which use code execution feedback to refine code iteratively. These include: Self-Editing~\cite{zhang2023self}, Self-Debugging (Trace)~\cite{chen2023teaching}, Self-Debugging (Explanation)~\cite{chen2023teaching}, LDB\cite{zhong2024ldb}, MGDebugger~\cite{shi2024code}, and Reflexion~\cite{shinn2023reflexion}.

\subsubsection{Implementation}
We select GPT-4o-mini~\cite{openai2024gpt4omini}, Claude-3.5-sonnet~\cite{anthropic2024claude}, and DeepSeek-Coder-V2-Lite~\cite{zhu2024deepseek} as the LLM backbones for our experiments. GPT-4o-mini and Claude-3.5-sonnet are accessed via their respective API forms, while DeepSeek-Coder-V2-Lite (16B) is locally deployed. Since all the methods utilize an iterative debugging process, we set the maximum debugging time to 5 iterations. All models are run with a temperature of 0. \footnote{The source code of this work is made available at \href{https://github.com/yevzh/NL-Debugging}{https://github.com/yevzh/NL-Debugging}.}




\subsection{Code Debugging Performance \textbf{(RQ1)}}

\begin{figure}[]
    \centering
    
    \includegraphics[width=\linewidth]{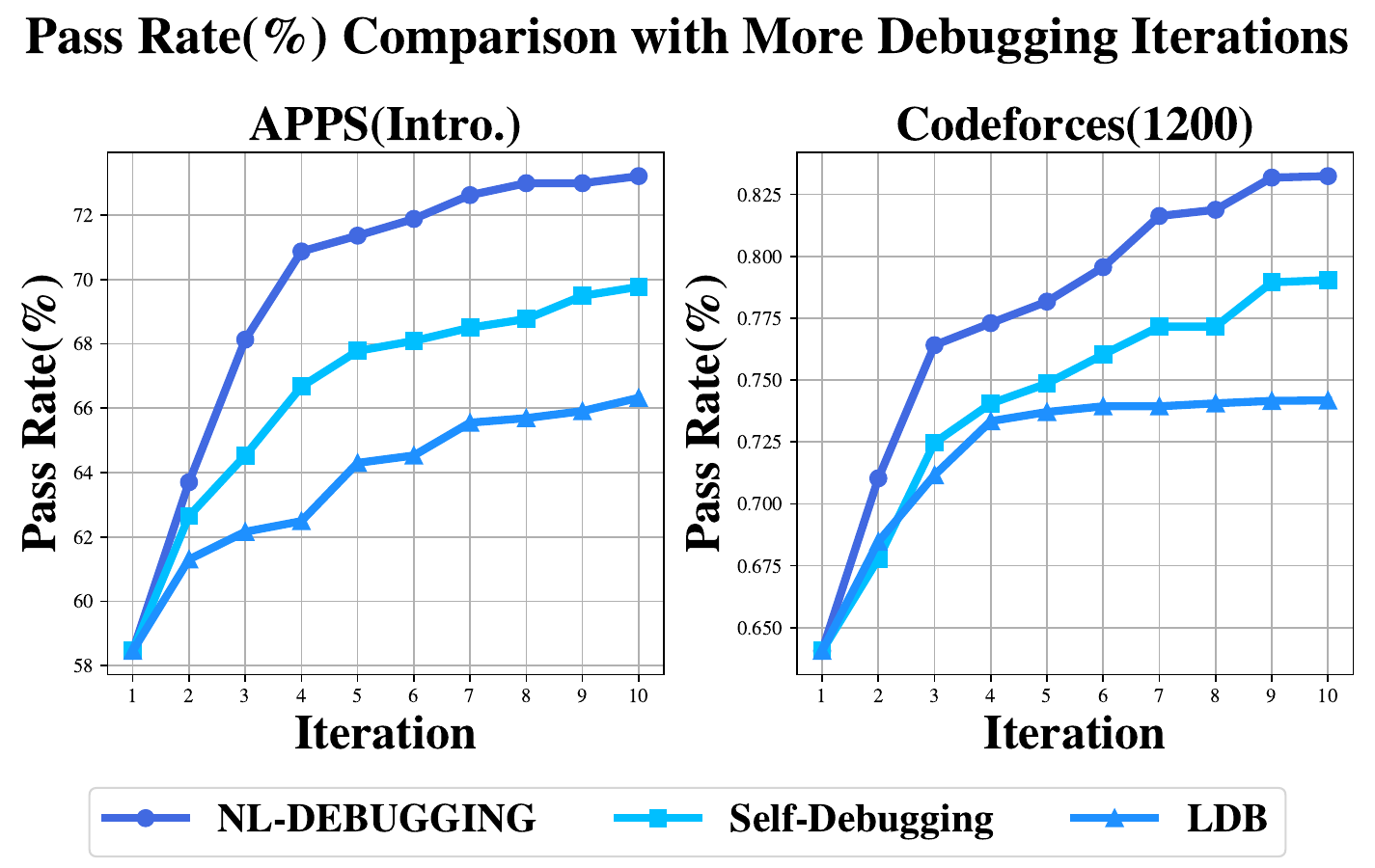}
    
    \caption{Pass rate (\%) comparison of \our{} and other code-level debugging methods with more debugging iterations.}
    \label{fig:iter}
\end{figure}


We compare the performance of \our{} and other code debugging methods in Table~\ref{tab:my-table}. Our findings are organized into three key observations:

\begin{itemize}[topsep = 3pt,leftmargin =10pt]
    \item \our{} shows significant improvements on contest-level programming tasks across all datasets, demonstrating its effectiveness in complex problem-solving scenarios where traditional code-level debugging struggles.

    \item Self-Debugging generally outperforms other baselines by reasoning directly at the code level. While methods like LDB and MGDB focus on detailed structural analysis of code components, they lack natural language optimization. \our{} advances beyond Self-Debugging by leveraging backtranslated natural language representations of buggy code, enabling a more global reasoning strategy than code-level stepwise refinements.

    \item At the APPS Introductory level, DeepSeek-Coder-V2-Lite shows minimal improvement in pass@1 compared to baseline methods. While its pass rate surpasses some baselines, its debugging gains are less significant than those of GPT-4o-mini and Claude3.5-Sonnet. This likely stems from DeepSeek-Coder-V2-Lite’s smaller parameter size and coder-focused design, which may limit its capacity for complex reasoning.

\end{itemize}

We further analyze how different debugging methods scale with the number of iterations, as shown in Figure~\ref{fig:iter}. Compared to competitive baselines, \our{} consistently maintains a performance advantage across iterations. Notably, as the number of iterations increases to 10, its performance improves steadily, demonstrating strong scalability and robustness in iterative refinement.

\subsection{What Format of Natural Language Representation Works the Best? \textbf{(RQ2)}}

\begin{table*}[]
\centering

\caption{Performance comparison of different types of natural language in \our{}.}
\label{tab:form}
\resizebox{\textwidth}{!}{%
\begin{tabular}{@{}clcccccccccccc@{}}
\toprule
\multirow{3}{*}{\textbf{Arch}} &
  \multirow{3}{*}{\textbf{NL Type}} &
  \multicolumn{6}{c}{\textbf{APPS}} &
  \multicolumn{6}{c}{\textbf{Codeforces}} \\
 &
   &
  \multicolumn{3}{c}{\textbf{Pass Rate(\%)}} &
  \multicolumn{3}{c}{\textbf{Pass@1(\%)}} &
  \multicolumn{3}{c}{\textbf{Pass Rate(\%)}} &
  \multicolumn{3}{c}{\textbf{Pass@1(\%)}} \\ \cmidrule(l){3-14} 
 &
   &
  \multicolumn{1}{l}{\textbf{Intro.}} &
  \multicolumn{1}{l}{\textbf{Inter.}} &
  \multicolumn{1}{l}{\textbf{Comp.}} &
  \multicolumn{1}{l}{\textbf{Intro.}} &
  \multicolumn{1}{l}{\textbf{Inter.}} &
  \multicolumn{1}{l}{\textbf{Comp.}} &
  \multicolumn{1}{l}{\textbf{1200}} &
  \multicolumn{1}{l}{\textbf{1500}} &
  \multicolumn{1}{l}{\textbf{1800}} &
  \multicolumn{1}{l}{\textbf{1200}} &
  \multicolumn{1}{l}{\textbf{1500}} &
  \multicolumn{1}{l}{\textbf{1800}} \\ \midrule
\multirow{3}{*}{\textbf{GPT-4o-mini}} &
  Pseudocode &
  65.32 &
  65.28 &
  40.50 &
  46 &
  42 &
  18 &
  70.22 &
  60.32 &
  43.32 &
  52 &
  32 &
  23 \\
 &
  Key points &
  68.97 &
  68.85 &
  40.67 &
  48 &
  44 &
  21 &
  75.27 &
  61.04 &
  43.98 &
  59 &
  37 &
  21 \\
  \rowcolor{Gray}
 &
  Sketch &
  \textbf{71.36} &
  \textbf{69.96} &
  \textbf{44.17} &
  \textbf{51} &
  \textbf{48} &
  \textbf{23} &
  \textbf{79.57} &
  \textbf{64.01} &
  \textbf{48.69} &
  \textbf{63} &
  \textbf{41} &
  \textbf{25} \\ \midrule
\multirow{3}{*}{\textbf{Claude3.5-Sonnet}} &
  Pseudocode &
  74.07 &
  73.97 &
  \textbf{59.83} &
  58 &
  58 &
  \textbf{45} &
  80.74 &
  68.02 &
  57.23 &
  71 &
  52 &
  43 \\
 &
  Key points &
  74.71 &
  74.74 &
  58.17 &
  58 &
  55 &
  41 &
  80.92 &
  70.72 &
  58.42 &
  72 &
  51 &
  45 \\
 \rowcolor{Gray}
 &
  Sketch &
  \textbf{77.98} &
  \textbf{76.72} &
  59.67 &
  \textbf{62} &
  \textbf{62} &
  44 &
  \textbf{85.82} &
  \textbf{74.16} &
  \textbf{61.28} &
  \textbf{79} &
  \textbf{58} &
  \textbf{47} \\ \midrule
\multirow{3}{*}{\textbf{DeepSeek-Coder-V2-Lite}} &
  Pseudocode &
  57.73 &
  50.62 &
  21.00 &
  33 &
  27 &
  6 &
  58.78 &
  45.02 &
  28.68 &
  39 &
  21 &
  10 \\
 &
  Key points &
  61.13 &
  \textbf{59.95} &
  26.67 &
  \textbf{40} &
  \textbf{32} &
  9 &
  63.30 &
  47.60 &
  29.95 &
  \textbf{49} &
  25 &
  11 \\
  \rowcolor{Gray}
 &
  Sketch &
  \textbf{62.23} &
  59.23 &
  \textbf{27.60} &
  \textbf{40} &
  \textbf{32} &
  \textbf{12} &
  \textbf{63.65} &
  \textbf{50.92} &
  \textbf{32.80} &
  48 &
  \textbf{27} &
  \textbf{12} \\ \bottomrule
\end{tabular}%
}
\end{table*}

To investigate which type of text is most effective for code debugging, we examine three natural language formats: pseudocode, key points, and sketches. We summarize the key features and characteristics of each type below, and the prompting details are presented in Appendix~\ref{pt:prompts}:

\begin{itemize}[topsep = 3pt,leftmargin =10pt]

\item \textbf{Pseudocode:} Pseudocode retains most of the structural elements of the code while incorporating natural language components to enhance clarity~\cite{wen2024unlocking}. It provides a high-level overview of the algorithm, making the logic more comprehensible by blending code-like syntax with descriptive elements.

\item \textbf {Key points:} This format abstracts the code’s main logic into concise key points, summarizing the key thought steps and operations~\cite{li2024rethinkmcts}. Key points are iteratively added and refined during debugging to provide a simplified, high-level view of the algorithm.

\item \textbf{Sketch:} The sketch format gives a more detailed natural language description, outlining the structure and specific implementation details~\cite{wang2024planning}. As debugging progresses, the description is refined for greater clarity and completeness, offering a flexible representation of the logic of the code.

\end{itemize}

The results in Table~\ref{tab:form} show that the Sketch format achieves the best overall performance. Compared to Pseudocode, Sketch offers more detailed natural language descriptions, making debugging more intuitive. Compared to Key points, Sketch provides a richer representation of code structure, enabling greater flexibility and a larger modification space, which supports more accurate natural language reasoning.

Moreover, for the DeepSeek-Coder-V2-Lite model, Key points and Sketch perform similarly, likely due to the model’s limited natural language reasoning ability. In this context, the simpler and more abstract Key points format is better suited for debugging. This suggests that when a model’s reasoning capacity is constrained, abstract natural language representations may be more effective as debugging aids.

\subsection{Why Natural Language Debugging Works? \textbf{(RQ3)}}
\subsubsection{NL2NL vs. NL2C}

\begin{figure*}[h]
    \centering
    
    \includegraphics[width=\linewidth]{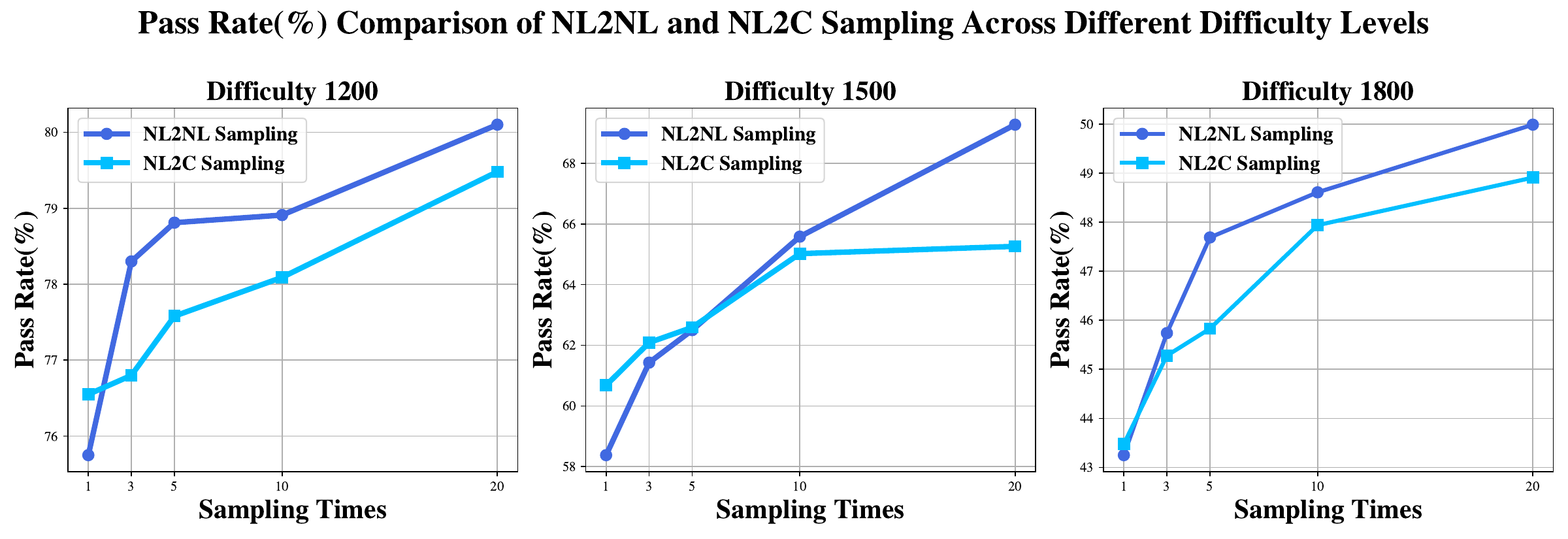}
    
    \caption{Pass rate (\%) comparison of NL2NL and NL2C sampling across different difficulty levels.}
    \label{fig:nl2nl_vs_nl2c}
\end{figure*}

\begin{table*}[]
\centering
\caption{Comparison of code-Level and \our{} in terms of structural and similarity metrics.}
\label{tab:code_vs_nl}
\resizebox{\textwidth}{!}{%
\begin{tabular}{@{}clccccccccc@{}}
\toprule
\multirow{2}{*}{\textbf{Dataset}} &
  \multicolumn{1}{c}{\multirow{2}{*}{\textbf{Method}}} &
  \multicolumn{3}{c}{\textbf{AST Edit Distance}} &
  \multicolumn{3}{c}{\textbf{Control Flow Depth Difference}} &
  \multicolumn{3}{c}{\textbf{BLEU Score}} \\ \cmidrule(l){3-11} 
 &
  \multicolumn{1}{c}{} &
  \textbf{Intro./1200} &
  \textbf{Inter./1500} &
  \textbf{Comp./1800} &
  \textbf{Intro./1200} &
  \textbf{Inter./1500} &
  \textbf{Comp./1800} &
  \textbf{Intro./1200} &
  \textbf{Inter./1500} &
  \textbf{Comp./1800} \\ \midrule
\multirow{2}{*}{\textbf{APPS}} &
  Code Level &
  3.03 &
  4.68 &
  2.86 &
  0.27 &
  0.23 &
  0.12 &
  0.8776 &
  0.8496 &
  0.9363 \\
  
 &
  \our{} &
  \textbf{5.57} &
  \textbf{7.31} &
  \textbf{4.94} &
  \textbf{0.33} &
  \textbf{0.37} &
  \textbf{0.25} &
  \textbf{0.7842} &
  \textbf{0.7318} &
  \textbf{0.8485} \\ \midrule
\multirow{2}{*}{\textbf{Codeforces}} &
  Code Level &
  2.19 &
  3.24 &
  3.34 &
  0.13 &
  0.18 &
  0.20 &
  0.9021 &
  0.8928 &
  0.8965 \\
 &
  \our{} &
  \textbf{5.25} &
  \textbf{6.15} &
  \textbf{5.87} &
  \textbf{0.21} &
  \textbf{0.37} &
  \textbf{0.31} &
  \textbf{0.8109} &
  \textbf{0.7767} &
  \textbf{0.8051} \\ \bottomrule
\end{tabular}%
}
\end{table*}

\begin{table}[]

\caption{Statistical results comparing modification distances and BLEU score between samples with pass rate growth greater than 0.5 and samples with no change.}
\label{tab:stat}
\resizebox{\columnwidth}{!}{%
\begin{tabular}{lccc}
\hline \toprule
\multicolumn{1}{c}{\textbf{}}                  & \textbf{Edit Distance} & \textbf{CFG Distance} & \textbf{BLEU Score} \\ \hline
Pass Rate Growth \textgreater 0.5 & 7.21                   & 0.57                                 & 0.6779              \\
No Change                         & 0.53                   & 0.05                                 & 0.9724              \\ \hline
\end{tabular}%
}
\end{table}

\our{} pipeline could be decomposed into two stages. The first stage, \textbf{NL2NL}, involves refining the initial natural language representation (i.e., the sketch) derived from the buggy code into an improved and corrected natural language form. The second stage, \textbf{NL2C}, refers to transforming this refined natural language representation into executable code. To analyze the relative importance of these stages, we employ multiple sampling attempts at each step and compare the performance gains they bring. 

Our experiments reveal that sampling during the NL2NL phase consistently results in greater improvements across all difficulty levels, as illustrated in Figure~\ref{fig:nl2nl_vs_nl2c}. This indicates that the performance gains are primarily driven by effective refinement in the natural language space, rather than by generating multiple candidate programs during the NL2C stage. In other words, the key to successful natural language debugging mainly lies in enhancing the quality of the underlying natural language abstraction.

\subsubsection{\our{} Enhances Diversity to Improve Debugging Performance}

To investigate natural language representations' benefits, we look into the specific change that \our{} brings. Table~\ref{tab:code_vs_nl} presents a comparative analysis between code-level debugging and \our{} across three key metrics to quantitatively evaluate the changes that debugging methods bring, following ~\citet{pan2025codecor, gong2024ast}: \textit{AST Edit Distance}, \textit{Control Flow Depth Difference}, and \textit{BLEU Score}. Detailed explanations of these metrics can be found in Appendix~\ref{app:metric_details}. Table~\ref{tab:stat} presents a statistical analysis comparing two sets of samples processed with code-level debugging: one with a significant pass rate improvement (greater than 0.5) and another where the pass rate did not change.

Our findings are organized into the following observations: 

\begin{itemize}[topsep = 3pt,leftmargin =10pt]
    \item \textbf{\our{} Brings More Diverse Changes:} Table~\ref{tab:code_vs_nl} shows significantly larger \textit{AST Edit Distance} values, higher \textit{Control Flow Depth Difference}, and lower \textit{BLEU Score} for \our{}, indicating more extensive and flexible modifications and more meaningful structural adjustments compared to code-level debugging.
    
    \item \textbf{Greater Diversity in Modifications Correlates with Better Debugging Performance:} As shown in Table~\ref{tab:stat}, samples with significant pass rate improvement exhibit notably higher \textit{Edit Distance} and \textit{Control Flow Depth Difference}, along with lower \textit{BLEU Score}, compared to samples where the pass rate did not change. This highlights the critical role of diversity in achieving effective debugging outcomes.
    
\end{itemize}

In summary, by introducing more diverse modifications, \our{} enhances debugging performance. These diverse changes enable more meaningful adjustments, especially for complex bugs, improving overall debugging effectiveness in addressing deeper logical or reasoning issues.
\subsection{How does Natural Language Work Better with Code Execution Feedback? \textbf{(RQ4)}}

\begin{figure}[h]

    \centering

    \includegraphics[width=\linewidth]{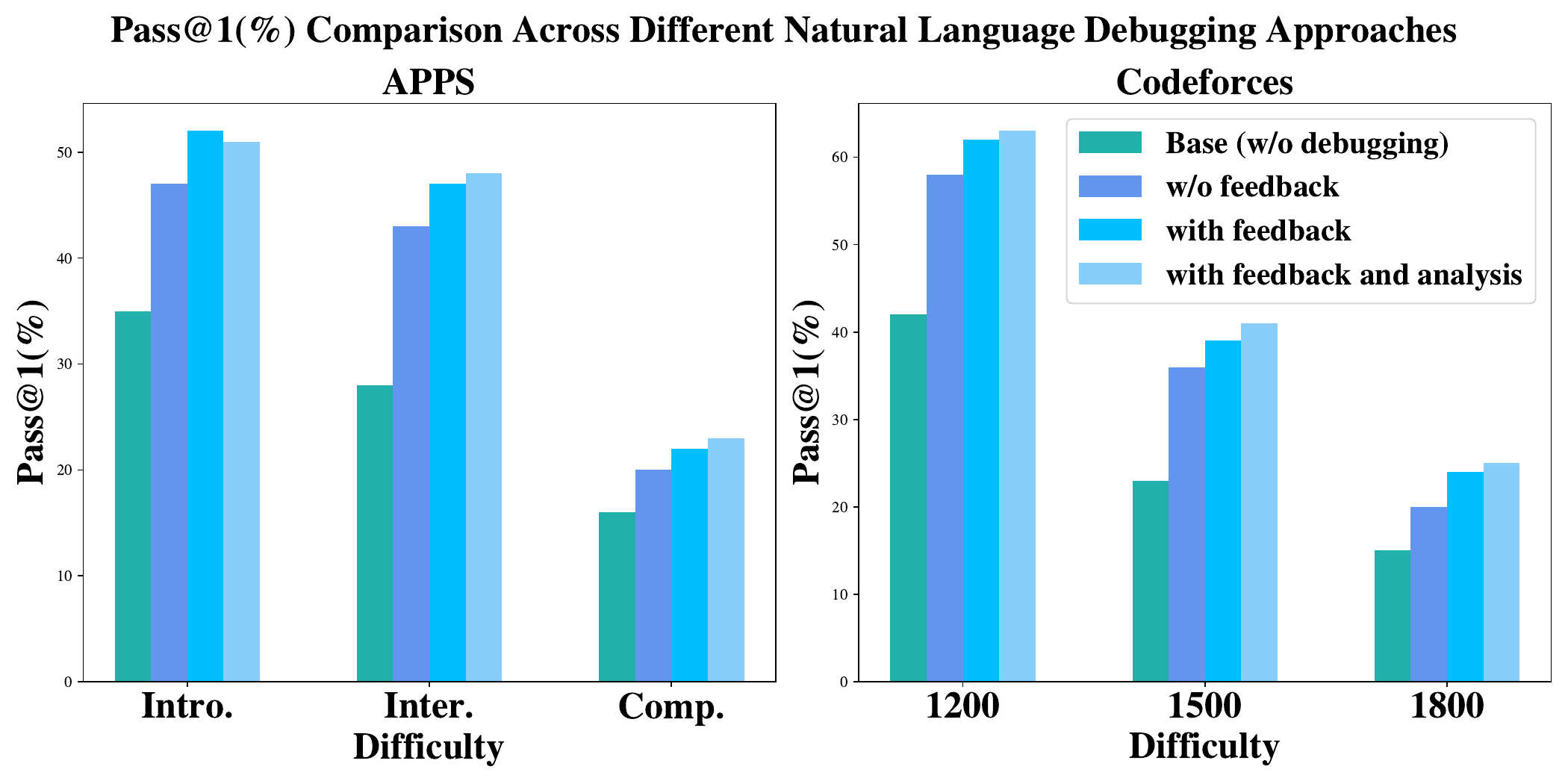}
    
    \caption{Pass@1 (\%) comparison across different natural language debugging approaches.} 
    \label{fig:feedback_analysis}
\end{figure}

\begin{table*}[]
\centering
\tiny
\caption{Performance comparison of \our{} and long CoT Approaches on APPS and Codeforces datasets.}
\label{tab:iterative_growth}
\resizebox{\textwidth}{!}{%
\begin{tabular}{@{}lcccccccccccc@{}}
\toprule
\multirow{3}{*}{\textbf{Method}} &
  \multicolumn{6}{c}{\textbf{APPS}} &
  \multicolumn{6}{c}{\textbf{Codeforces}} \\
 &
  \multicolumn{3}{c}{\textbf{Pass Rate(\%)}} &
  \multicolumn{3}{c}{\textbf{Pass@1(\%)}} &
  \multicolumn{3}{c}{\textbf{Pass Rate(\%)}} &
  \multicolumn{3}{c}{\textbf{Pass@1(\%)}} \\ \cmidrule(l){2-13} 
 &
  \multicolumn{1}{l}{\textbf{Intro.}} &
  \multicolumn{1}{l}{\textbf{Inter.}} &
  \multicolumn{1}{l}{\textbf{Comp.}} &
  \multicolumn{1}{l}{\textbf{Intro.}} &
  \multicolumn{1}{l}{\textbf{Inter.}} &
  \multicolumn{1}{l}{\textbf{Comp.}} &
  \multicolumn{1}{l}{\textbf{1200}} &
  \multicolumn{1}{l}{\textbf{1500}} &
  \multicolumn{1}{l}{\textbf{1800}} &
  \multicolumn{1}{l}{\textbf{1200}} &
  \multicolumn{1}{l}{\textbf{1500}} &
  \multicolumn{1}{l}{\textbf{1800}} \\ \midrule
\our{} &
  \textbf{71.36} &
  \textbf{69.96} &
  44.17 &
  \textbf{51} &
  \textbf{48} &
  23 &
  \textbf{79.57} &
  \textbf{64.00} &
  \textbf{48.69} &
  \textbf{65} &
  \textbf{41} &
  \textbf{25} \\
Long CoT(Direct) &
  67.05 &
  63.73 &
  42.50 &
  48 &
  40 &
  23 &
  74.50 &
  62.14 &
  43.79 &
  55 &
  37 &
  21 \\
Long CoT(Sketch) &
  71.08 &
  68.16 &
  \textbf{45.17} &
  \textbf{51} &
  47 &
  \textbf{26} &
  77.36 &
  63.96 &
  45.12 &
  60 &
  38 &
  22 \\ \bottomrule
\end{tabular}%
}
\end{table*}

To investigate how natural language can work better for debugging, we focus on the impact of execution feedback within \our{}. Figure~\ref{fig:feedback_analysis} compares pass@1 performance across different debugging strategies on the APPS and Codeforces benchmarks, with varying feedback mechanisms.

The results show that incorporating execution feedback significantly improves debugging performance, confirming its importance in natural language-based debugging. Specifically, execution feedback information is crucial, providing the foundation for effective reasoning. 

Additionally, adding an analysis step to raw execution feedback leads to further performance improvements. This analysis step, akin to \textit{self-debugging} as opposed to \textit{self-editing}, helps the model reason more deeply about the code, underscoring the value of deeper reasoning for more accurate debugging.


\subsection{Is O1-like Long-CoT Beneficial for \our{}? \textbf{(RQ5)}}


Inspired by o1-like reasoning methods~\cite{jaech2024openai, qin2024o1, guo2025deepseek}, which incorporate reflection and iterative refinement of thought chains, we investigated whether similar approaches could enhance debugging in natural language. In the context of \our{}, where execution feedback is available, we aimed to emulate this reflective reasoning process by incorporating feedback to allow the model to pause, analyze, and refine its reasoning.
To assess the effectiveness of long chains of thought in debugging, we propose two approaches:
\begin{itemize}[topsep = 3pt,leftmargin =10pt]
    \item \textbf{Lont CoT (Direct):} This approach simulates iterative reasoning by analyzing execution feedback and appending insights to an ongoing reasoning chain, incrementally refining the model's understanding.
    
    \item \textbf{Lont CoT (Sketch):} This method uses the iteratively grown chain of thought from the first approach as a basis for generating structured natural language sketches. The goal is to stimulate critical thinking in sketch refinement and develop more comprehensive problem-solving strategies.
\end{itemize}

Based on the results in Table~\ref{tab:iterative_growth}, we make the following observations:

\begin{itemize}[topsep = 3pt,leftmargin =10pt]
     \item \textbf{Long CoT Approaches Underperform \our{}:} Simply appending iterative trial-and-error reflections does not improve debugging effectiveness. The longer reasoning chains fail to consistently enhance performance, indicating that added context alone is insufficient.

    \item \textbf{Long CoT Structure is Less Effective than Sketch as Intermediary for Code Regeneration:} Directly appending execution feedback to Long CoT (Long CoT Direct) yields worse results than using these chains to produce a coherent sketch (Long CoT Sketch). This suggests that the natural language sketch format better supports code generation, while manually appended Long CoT lacks the necessary coherence.
\end{itemize}

In summary, manual appending of reasoning chains fails to stimulate LLMs’ critical self-reflection reliably without parameter tuning. Therefore, \our{}’s focus on direct refinement of natural language sketches proves more effective for debugging.

\section{Conclusion}
In this paper, we introduce \our{}, which leverages natural language as an intermediate representation to enhance code debugging efficiency and accuracy significantly. Experimental results show that \our{} outperforms traditional code-level debugging, especially in addressing deep algorithmic flaws. Our results suggest that using natural language sketches is the most effective format for code debugging. Further, we demonstrate that \our{} facilitates a broader modification space, enabling more diverse corrections and more effective debugging performance. Overall, \our{} demonstrates the important potential of natural language reasoning in code debugging, offering a promising direction for future research in automated debugging.

\section*{Acknowledgements}
The Shanghai Jiao Tong University team is partially supported by National Key R\&D Program of China (2022ZD0114804), Shanghai Municipal Science and Technology Major Project (2021SHZDZX0102) and National Natural Science Foundation of China (62322603, 62177033, 62502310).

\section*{Limitations}

\paragraph{Expanding Modification Space through Tree Search} While our approach utilizes iterative refinement for natural language representations, enhancing the modification space could be achieved by integrating tree search mechanisms into the reasoning process. By implementing tree search strategies, we could facilitate a more structured exploration of potential modifications based on natural language, allowing for a broader range of solutions.

\paragraph{Integrating Debugging Methods at Different Granularities} Integrating debugging methods at different granularities presents an opportunity for further enhancement. While code-level debugging has its advantages, natural language reasoning offers unique strengths. Combining analyses at various levels may facilitate a more effective resolution of unknown bugs by leveraging the strengths of both approaches, potentially leading to improved debugging outcomes.

\paragraph{Further Exploring Long Chains of Thought} Our experiments revealed that manually constructed trial-and-error reasoning processes did not yield successful outcomes. However, this indicates a limitation in our current approach rather than a dismissal of long chains of thought. Methods like journey learning, as demonstrated in the O1-Journey project~\cite{qin2024o1}, suggest that LLMs can be trained with relatively few samples to produce coherent long reasoning chains through exploration and reflection. Applying these insights to the code domain could enhance our understanding of effectively implementing long chains of thought in debugging scenarios.

\paragraph{Generalizing to Software Reasoning Tasks} Our current exploration primarily focuses on debugging algorithmic programming challenges with execution feedback. While this focus has allowed us to address specific code generation practices, the methodologies and insights derived from this work could generalize to other software reasoning tasks. For instance, applying our framework to debugging software issues in development environments or addressing problems in legacy codebases~\cite{jimenez2023swe} could yield significant benefits, highlighting the broader applicability of natural language reasoning in diverse programming contexts.


\bibliography{custom}
\newpage
\appendix
\section*{Appendix}
\section{Dataset Details}
We evaluate our \our{} framework on two widely used code datasets: APPS~\cite{hendrycks2021measuring} and Codeforces~\citep{codeforcesweb}. The APPS dataset includes three difficulty levels—introductory, interview, and competition—with a total of 5000 programming problems for training and 5000 for testing. The Codeforces dataset contains problems from the Codeforces online programming contest, with varying difficulty levels categorized by "ratings", from which we choose 1200, 1500, and 1800 for evaluation. 

For both datasets, we select 100 problems from each difficulty level for evaluation to ensure a balanced assessment. Both datasets use the same set of test cases for algorithm optimization (public test cases) and performance evaluation (private test cases). The public test cases are used during the algorithm running, while the private test cases are reserved for evaluating the generated codes.

\section{Baseline Details}
\label{app:baseline_details}
To provide a comprehensive understanding of our experimental setup, we present a detailed overview of each baseline code debugging method used for comparison with \our{}. All baselines leverage code execution feedback to refine code iteratively.

\paragraph{Self-Edit~\cite{zhang2023self}} This method utilizes the execution results from test cases to regenerate code, aiming to correct errors based on observed failures. The model iteratively edits code based on execution feedback but does not explicitly reason beyond localized fixes.

\paragraph{Self-Debug(Explanation)~\cite{chen2023teaching}} This method augments the code-level analysis by prompting the model to generate natural language explanations of its code. While it introduces a reasoning step, the process still centers on interpreting and modifying the code.

\paragraph{Self-Debug(Trace)~\cite{chen2023teaching}} This variant performs execution tracing to follow the flow of control and variable states. The reasoning it enables is constrained to analyzing runtime behavior within the original program structure.

\paragraph{LDB~\cite{zhong2024ldb}} LDB decomposes code into basic blocks and performs execution analysis on each block. Its reasoning is fine-grained and structural, but strictly within the code block domain.

\paragraph{MGDebugger~\cite{shi2024code}} This method builds a hierarchical tree of subfunctions and performs bottom-up debugging. The reasoning is hierarchical but still grounded in the syntactic and structural composition of the buggy code.

\paragraph{Reflexion~\cite{shinn2023reflexion}} Reflexion leverages past execution traces to guide future code modifications. Although it reflects on previous failures, the reasoning process remains code-centric and operates directly on the source program.

In contrast to baseline methods that perform reasoning directly on buggy code, \our{} distinguishes itself by conducting reasoning on a natural language abstraction of the code. This shift allows \our{} to operate at a higher level of semantic understanding, facilitating more global and flexible reasoning about the algorithm’s intent rather than being confined to localized code-level analysis.

We note that \our{} adopts a multi-round iterative refinement process, consistent with baselines like Self-Debugging~\cite{chen2023teaching} and LDB~\cite{zhong2024ldb} to ensure fairness. At each iteration, \our{} refines the natural language representation using execution feedback, regenerates code, and continues this process until a solution passes all visible test cases or reaches a maximum iteration limit.

\begin{figure*}[h]
    \centering
    \tiny
    \includegraphics[width=\linewidth]{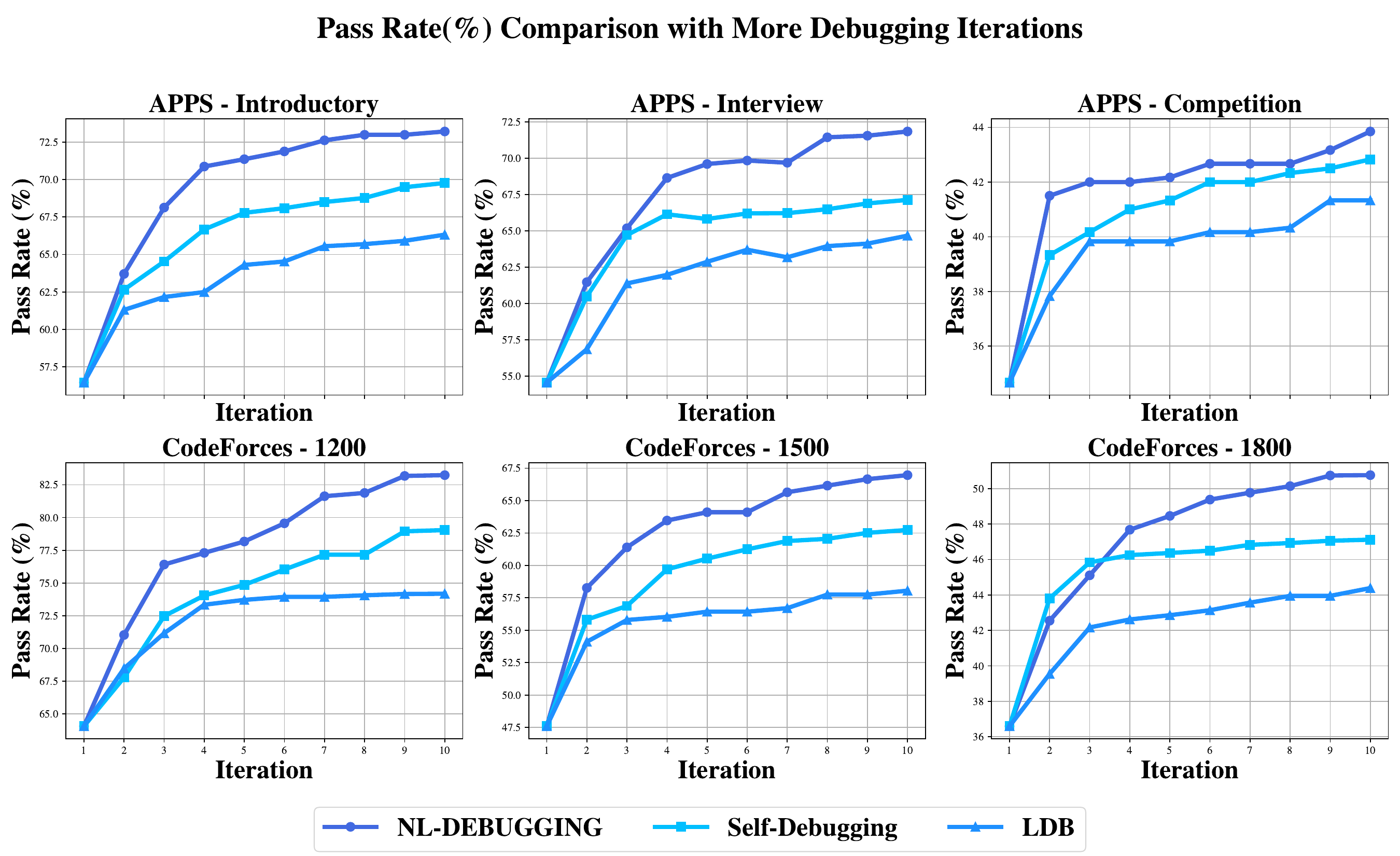}
    
    \caption{Pass rate (\%) comparison of \our{} and other code debugging methods with more debugging iterations on both datasets.}
    \label{fig:full_iter}
\end{figure*}
\section{Metric Details}
\label{app:metric_details}
To provide a clear understanding of how we assessed the impact of \our{}, we provide a description of each metric that is evaluated.

\subsection*{AST Edit Distance}
AST Edit Distance measures the structural difference between the original code's Abstract Syntax Trees (ASTs) and the debugged code~\cite{pan2025codecor}. A higher AST Edit Distance indicates that the debugging process has resulted in more significant structural modifications to the code, suggesting more extensive changes to the code's implementation. It is calculated as the number of edits (insertions, deletions, and substitutions) required to transform one AST into another.

\subsection*{Control Flow Depth Difference}
Control Flow Depth Difference quantifies the change in the depth of the control flow within the code~\cite{gong2024ast}. It is calculated as the absolute difference between the original code's control flow graph (CFG) 's maximum depth and the CFG of the debugged code. A higher Control Flow Depth Difference suggests that debugging leads to more significant alterations in the code's control structure.

\subsection*{BLEU Score}
BLEU (Bilingual Evaluation Understudy) Score is a metric used to evaluate the similarity between the original and debugged code~\cite{papineni2002bleu}. It measures the n-gram overlap between the two code snippets, with higher scores indicating greater similarity. In our analysis, a lower BLEU score suggests that \our{} results in more substantial modifications, leading to more flexible and impactful error corrections.

\begin{figure*}[h]
    \centering
    
    \includegraphics[width=\linewidth]{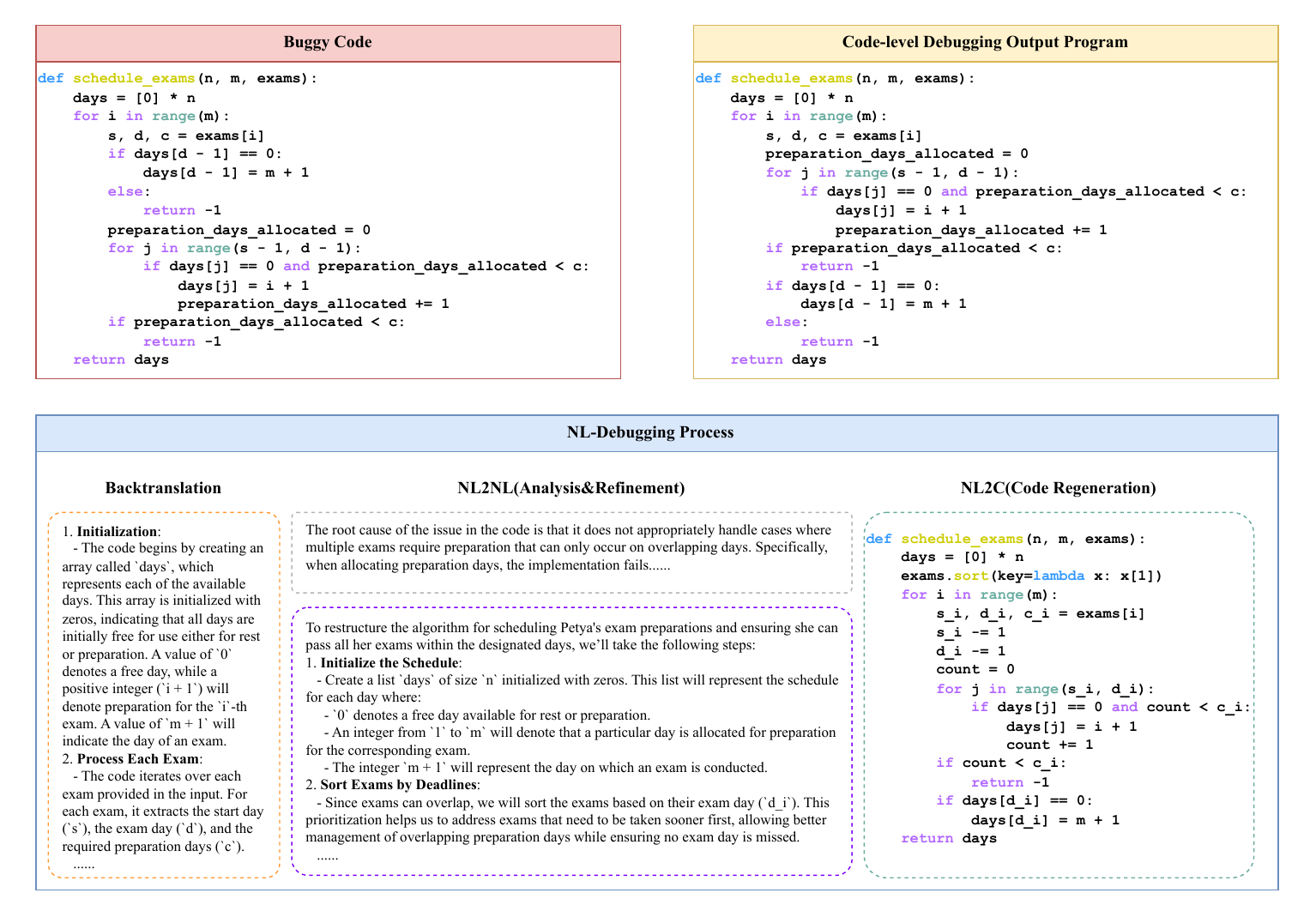}
    
    \caption{A case study for \our{}.}
    \label{fig:case}
\end{figure*}

\section{More Experimental Results}

In this section, we present the complete experimental results comparing the performance of \our{} with two code-level debugging approaches. Figure~\ref{fig:full_iter} provides a comprehensive analysis across three difficulty levels on two datasets, illustrating the robustness and consistency of our method.

The experimental findings yield the following conclusions:

\begin{itemize}[topsep = 3pt,leftmargin =10pt]
    \item \our{} consistently outperforms code-level debugging approaches at various iteration stages, demonstrating its robustness and scalability.

    \item With increasing iterations, \our{} exhibits a sustained improvement in performance, maintaining a consistent lead over competing approaches in both datasets. This highlights its ability to adapt and enhance debugging outcomes over iterative refinements.
\end{itemize}

\section{Case Study}

This section presents a case study illustrating the effectiveness of \our{} in addressing logical errors. The case focuses on a buggy code snippet (Figure~\ref{fig:case}) that fails to prioritize exams by their deadlines, leading to potential scheduling conflicts. Sorting exams according to their deadlines was identified as a key solution. Self-Debugging~\cite{chen2023teaching}, a representative code-level debugging approach, attempts to fix local issues such as marking exam days and allocating preparation time. However, it fails to detect the root cause—the absence of a prioritization mechanism for handling overlapping preparation days among multiple exams. This highlights the limitations of code-level debugging in resolving complex logical errors.

In contrast, \our{} adopts a systematic approach. During the Sketch refinement phase, natural language reasoning reveals the missing prioritization logic responsible for scheduling conflicts. By regenerating code based on the corrected sketch and successfully passing test cases, \our{} demonstrates its ability to address deeper logical flaws. This case exemplifies how \our{} surpasses traditional code-level debugging by identifying and resolving fundamental logical inconsistencies often overlooked by conventional methods.

\begin{table*}[]

\caption{Distribution of logical and runtime errors in LLM-generated code across different  models.}
\label{tab:count}
\resizebox{\textwidth}{!}{%
\begin{tabular}{l|llllll}
\hline \toprule
\multicolumn{1}{c|}{\textbf{Model}} & \multicolumn{1}{c}{\textbf{APPS(Intro.)}} & \multicolumn{1}{c}{\textbf{APPS(Inter.)}} & \multicolumn{1}{c}{\textbf{APPS(Comp.)}} & \multicolumn{1}{c}{\textbf{CodeForce-1200}} & \multicolumn{1}{c}{\textbf{CodeForce-1500}} & \multicolumn{1}{c}{\textbf{CodeForce-1800}} \\  \midrule
GPT-4o-mini                         & 100\% Logical / 0\% Runtime               & 100\% Logical / 0\% Runtime               & 100\% Logical / 0\% Runtime              & 100\% Logical / 0\% Runtime                 & 99\% Logical / 1\% Runtime                  & 100\% Logical / 0\% Runtime                 \\
GPT-3.5-turbo                       & 100\% Logical / 0\% Runtime               & 100\% Logical / 0\% Runtime               & 98\% Logical / 2\% Runtime               & 100\% Logical / 0\% Runtime                 & 100\% Logical / 0\% Runtime                 & 100\% Logical / 0\% Runtime                 \\
DeepSeek-Coder-V2-Lite              & 100\% Logical / 0\% Runtime               & 100\% Logical / 0\% Runtime               & 99\% Logical / 1\% Runtime               & 100\% Logical / 0\% Runtime                 & 99\% Logical / 1\% Runtime                  & 100\% Logical / 0\% Runtime                 \\ \bottomrule
\end{tabular}%
}
\end{table*}

\section{More Details on Experimental Settings}
To further clarify our experimental setting and methodological scope, this section provides additional context regarding the relationship between debugging and code generation and the distinction between our work and traditional automated program repair (APR).

\subsection{Code Debugging vs. Code Generation: Their Connection}
Code debugging is a specific step within the broader code generation process. The common paradigm in code generation follows the sequence: problem → seed program → debugged program. Our approach focuses specifically on transforming the seed program into the final program. Both debugging and code generation use the same evaluation metrics: pass rate and pass@1. While the settings for debugging and code generation differ, the key distinction is that debugging starts from a fixed seed program for all methods, where analysis and reasoning are performed to correct the buggy code. The pass rate improvement directly reflects the debugging process's effectiveness, demonstrating the ability to generate a fixed version of the buggy code.


\subsection{Code Debugging vs. Automatic Program Repair: Their Difference}
Automated Program Repair (APR) methods typically focus on repairing well-defined errors, such as syntax or semantic issues, that prevent code from executing correctly. These methods work with curated bug benchmarks, generating patches based on predefined error patterns. APR mainly addresses execution-related faults, such as missing return statements or incorrect variable initialization.

In contrast, our Code Debugging setting targets LLM-generated code, which often contains logical rather than runtime errors. These errors occur when the code logic fails to align with the intended functionality, even if the code is syntactically correct and executes without crashing. As shown in Table~\ref{tab:count}, LLM-generated code predominantly contains logical errors, with very few runtime errors. This emphasizes our approach's focus on correcting logical flaws at the natural language level, rather than dealing with execution failures.

\section{Prompts}
\label{pt:prompts} %

\subsection{Prompts for \our{}}

\label{app:nldb}
Tables \ref{tab:prompt_back}, \ref{tab:prompt_exe}, \ref{tab:prompt_refine}, and \ref{tab:prompt_regen} present prompts that represent the entire framework’s operation process, including steps for backtranslation, execution analysis, refinement, and code regeneration. In this framework, we adopt the \textbf{Sketch} format as the default natural language representation of the program throughout the pipeline.

\begin{table*}[h!]
\centering
\resizebox{0.98\textwidth}{!}{
\begin{tcolorbox}[colback=blue!5!white,colframe=black!55!black,width=0.98\textwidth,title={Prompt for Backtranslation}]
\small
You are an expert Python programmer. Below is an algorithmic question (problem specification) along with the current implementation for solving the problem.\\
$\{\text{Problem Description}\}$\\
$\{\text{Current Code Implementation}\}$\\
Your task is to generate a **Natural Language Sketch** for this code.\\
This sketch should describe the logical reasoning or steps that the code is trying to follow in order to solve the problem.\\
Do not focus on syntax or specific code lines, but explain the thought process or approach the code takes to solve the problem at a high level.\\

\end{tcolorbox}
}
\caption{Prompt for Basktranslation.}
\label{tab:prompt_back}
\end{table*}

\begin{table*}[h!]
\centering
\resizebox{0.98\textwidth}{!}{
\begin{tcolorbox}[colback=blue!5!white,colframe=black!55!black,width=0.98\textwidth,title={Prompt for Execution Analysis}]
\small
You are an expert Python programmer. You will be provided with an algorithmic problem description, the current Python code implementation, and the execution feedback that indicates where the code went wrong.\\
$\{\text{Problem Description}\}$\\
$\{\text{Current Code Implementation}\}$\\
$\{\text{Execution Feedback}\}$\\
$\{\text{Natural Language Sketch with Bugs}\}$\\
Please analyze the feedback and provide an explanation of what went wrong in the code and why it failed in this sketch.\\
Do not provide specific steps to fix the sketch. Focus solely on explaining the root cause of the issue in two or three sentences.\\
\end{tcolorbox}
}
\caption{Prompt for Execution Analysis.}
\label{tab:prompt_exe}
\end{table*}

\begin{table*}[h!]
\centering
\resizebox{0.98\textwidth}{!}{
\begin{tcolorbox}[colback=blue!5!white,colframe=black!55!black,width=0.98\textwidth,title={Prompt for Natural Language Refinement}]
\small
You are an expert Python programmer. Below is an algorithmic problem description, the current natural language sketch of the solution (which contains bugs), the current code implementation, and the feedback from running the code, as well as a detailed expert analysis of the bug.\\
$\{\text{Problem Description}\}$\\
$\{\text{Natural Language Sketch with Bugs}\}$\\
$\{\text{Current Code Implementation}\}$\\
$\{\text{Bug Analysis}\}$\\
Based on the feedback and expert analysis of the current sketch, please provide a refined and corrected version of the sketch. The corrected sketch should:\\
1. Identify and correct specific points in the current sketch where errors or incorrect assumptions have been made.\\
2. Expand and elaborate on problematic steps in greater detail to explain the correct reasoning, ensuring that you address and fix the issues identified in the previous sketch.\\
3. Ensure that each step is logically connected and that any potential issues in the original approach are explicitly avoided in the new version.\\
Do not include any specific code. The goal is to refine and improve the high-level natural language explanation of the problem-solving approach.\\

\end{tcolorbox}
}
\caption{Prompt for Natural Language Refinement.}
\label{tab:prompt_refine}
\end{table*}

\begin{table*}[h!]
\centering
\resizebox{0.98\textwidth}{!}{
\begin{tcolorbox}[colback=blue!5!white,colframe=black!55!black,width=0.98\textwidth,title={Prompt for Regeneration}]
\small
You are an expert Python programmer. Below is a high-level natural language sketch for a correct solution to an algorithmic problem, along with the problem description.\\
$\{\text{Problem Description}\}$\\
$\{\text{Refined Natural Language Sketch}\}$\\
Your task is to write Python code that implements this sketch and solves the problem.\\
Do not include unnecessary comments or explanations, only the code itself.\\
\end{tcolorbox}
}
\caption{Prompt for Regeneration.}
\label{tab:prompt_regen}
\end{table*}

\subsection{Prompts for Different Types of Natural Language}

\label{app:different}

Tables~\ref{tab:keypoints}, \ref{tab:keypoints2}, \ref{tab:pesudocode}, and \ref{tab:pseudocode2} define the backtranslation and refinement prompts for \textbf{key points} and \textbf{pseudocode} generation. The corresponding prompts for the \textbf{Sketch} format, which serves as the default representation in our framework, are provided in Appendix~\ref{app:nldb}.

\begin{table*}[h!]
\centering
\resizebox{0.98\textwidth}{!}{
\begin{tcolorbox}[colback=blue!5!white,colframe=black!55!black,width=0.98\textwidth,title={Prompt for Key Points Backtranslation}]
\small
You are an expert Python programmer. Below is an algorithmic question (problem specification) along with the current implementation for solving the problem.\\
$\{\text{Problem Description}\}$\\
$\{\text{Current Code Implementation}\}$\\
I need you to extract 3 key points (thoughts) that summarize the core algorithm or logic used in this code.\\
Please list each thought in a separate entry in the format:\\
The output should be a **list of dicts** with each key as `Thought-i`. Do not include explanations or justifications, just focus on capturing the key algorithmic points.\\
\text{[}\\
    \{"Thought-1": "We could use the print function to finish the task in one line: print(2 + 3)."\},\\
    \{"Thought-2": "We should calculate the problem by setting a=2+3, and then print(a)."\},\\
    \{"Thought-3": "The problem can't be solved by Python."\} \\
\text{]}
\end{tcolorbox}
}
\caption{Prompt for Key Points Refinement.}
\label{tab:keypoints}
\end{table*}

\begin{table*}[h!]
\centering
\resizebox{0.98\textwidth}{!}{
\begin{tcolorbox}[colback=blue!5!white,colframe=black!55!black,width=0.98\textwidth,title={Prompt for Key Points Refinement}]
\small
You are an expert Python programmer. Below is an algorithmic problem description, the current list of thoughts (which contains bugs), the current code implementation, and the feedback from running the code.\\
$\{\text{Problem Description}\}$\\
$\{\text{Key Points with Bugs}\}$\\
$\{\text{Current Code Implementation}\}$\\
$\{\text{Bug Analysis}\}$\\
Please analyze and generate 1-2 new thoughts to correct the approach.\\
**Format Requirements:**\\
1. Output ONLY a JSON-parsable list of dictionaries\\
2. Each dictionary must use 'Thought-i' as key (i continues from previous sequence)\\
3. Each value should concisely state one algorithmic insight.\\
3. New thoughts can correct the previous buggy thoughts.\\
\text{[}\\
    \{"Thought-4": "Using depth-first search for tree traversal"\},\\
    \{"Thought-5": "Handling leaf nodes with null checks"\},\\
\text{]}
\end{tcolorbox}
}
\caption{Prompt for Key Points Refinement.}
\label{tab:keypoints2}
\end{table*}

\begin{table*}[h!]
\centering
\resizebox{0.98\textwidth}{!}{
\begin{tcolorbox}[colback=blue!5!white,colframe=black!55!black,width=0.98\textwidth,title={Prompt for Pseudocode Backtranslation}]
\small
You are an expert Python programmer. Below is an algorithmic question (problem specification) along with the current implementation for solving the problem.\\
$\{\text{Problem Description}\}$\\
$\{\text{Current Code Implementation}\}$\\
Your task is to generate **pseudocode** for this code in LaTeX format using the algorithm2e package.\\
This pseudocode should provide the high-level logic of the code. Format it as LaTeX pseudocode that captures the main steps and logic of the code.
\end{tcolorbox}
}
\caption{Prompt for Pseudocode Backtranslation.}
\label{tab:pesudocode}
\end{table*}

\begin{table*}[h!]
\centering
\resizebox{0.98\textwidth}{!}{
\begin{tcolorbox}[colback=blue!5!white,colframe=black!55!black,width=0.98\textwidth,title={Prompt for Pseudocode Refinement}]
\small
You are an expert Python programmer. Below is an algorithmic problem description, the current pseudocode of the solution (which contains bugs), the current code implementation, and the feedback from running the code.
$\{\text{Problem Description}\}$\\
$\{\text{Pseudocode with Bugs}\}$\\
$\{\text{Current Code Implementation}\}$\\
$\{\text{Bug Analysis}\}$\\
Please rethink the approach and generate a new, corrected pseudocode for solving the problem referring to the bug explanation for this buggy pseudocode.\\
This new pseudocode should address the issues in the previous pseudocode, focus on outlining the corrected steps needed to solve the problem, and avoid the errors found in the previous pseudocode."
\end{tcolorbox}
}
\caption{Prompt for Pseudocode Refinement.}
\label{tab:pseudocode2}
\end{table*}

\end{document}